\def\year{2018}\relax
\documentclass[letterpaper]{article} 
\usepackage{aaai18}  
\usepackage{times}  
\usepackage{helvet}  
\usepackage{courier}  
\usepackage{url}  
\usepackage{graphicx}  
\usepackage{amsfonts}       
\usepackage{amssymb}
\usepackage{amsmath}
\usepackage{subcaption}
\begin{document}
%
\title{No Modes left behind: Capturing the data distribution effectively using GANs}

\author{Shashank Sharma \and Vinay P. Namboodiri\\
Dept. of Computer Science and Engineering\\
Indian Institute of Technology, Kanpur\\
}
\maketitle
\begin{abstract}
Generative adversarial networks (GANs) while being very versatile in realistic image synthesis, still are sensitive to the input distribution. Given a set of data that has an imbalance in the distribution, the networks are susceptible to missing modes and not capturing the data distribution. While various methods have been tried to improve training of GANs, these have not addressed the challenges of covering the full data distribution. Specifically, a generator is not penalized for missing a mode. We show that these are therefore still susceptible to not capturing the full data distribution.

In this paper, we propose a simple approach that combines an encoder based objective with novel loss functions for generator and discriminator that improves the solution in terms of capturing missing modes. We validate that the proposed method results in substantial improvements through its detailed analysis on toy and real datasets. The quantitative and qualitative results demonstrate that the proposed method improves the solution for the problem of missing modes and improves training of GANs.

\end{abstract}

\section{Introduction}

\begin{figure}
    \centering
    \includegraphics[width=0.45\textwidth]{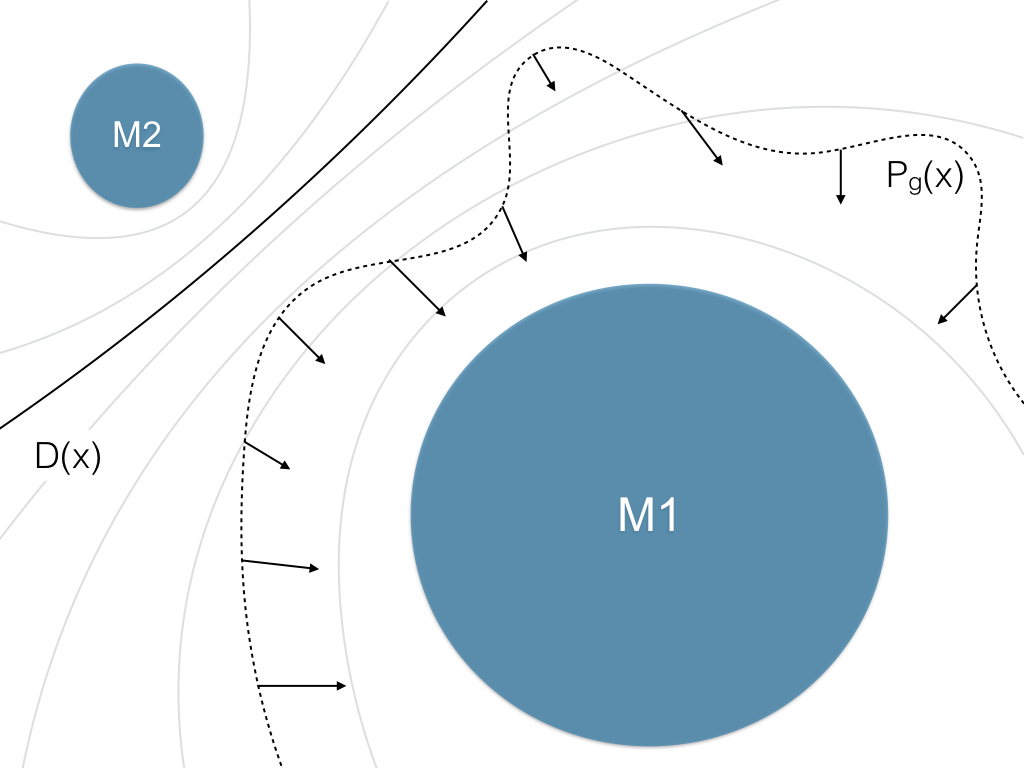}
    \caption{Shows how the Discriminator follows the local gradients to lose the smaller mode M2 without any penalty}
    \label{fig:mode_loss}
\end{figure}

Generative Adversarial Networks \cite{gan} aim to learn complex data distributions by learning a mapping from a noise distribution (Uniform/Gaussian) to a distribution that is based on an unsupervised set of samples. This task, while challenging has recently seen a number of successes.
The method learns to solve this problem by using a pair of networks a Generator \(G(z)\) and a Discriminator \(D(x)\), that are simultaneously trained by playing a minimax game. The Discriminator tries to differentiate between the real and the generated samples, and the Generator tries to fool the Discriminator by generating samples that are close to the real data. The training continues till \(G\) and \(D\) attain a Nash Equilibrium. Initially, GANs used to be notoriously hard to train but recent developments, \cite{improved_gan,wgan,iwgan,f_gan,unrolled_gan,obj_gan} have proposed modifications to the vanilla method that have resulted in more stable training methods. However, the methods still have limitations in terms of missing modes of data that are hard to capture as they may have a fewer number of samples. This problem is illustrated in the figure~\ref{fig:mode_loss} that shows how a discriminator is attracted towards its dominant mode and misses out on the smaller mode.

There have been works to encourage diversity in the generations, \cite{madgan}, but none to our knowledge exist that attempt to tackle the problem directly. Using the approach mentioned in BIGAN \cite{bigan} as our base objective, we propose a modification that addresses the mentioned problem in a much better way.

We test our model extensively against natural images from popular datasets like Cifar10 \cite{cifar}, CelebA \cite{celeba}; and an unusual dataset, frames from a surveillance video, \cite{traffic}. Since the background remains the same in all the frames, the dataset is highly clustered and encourages mode collapse. While the vanilla objective fails and collapses to a single mode, our method is able to train fairly well on it. Showing that unless the objective is not modeled for mode capturing, it fails.

\section{Background}

\subsection{Encoder}
Recently the use of a complementary network, an encoder, has become popular. It is usually trained to give the best latent representation of the data point, or inverting the generator $G$. Thus,
\[\exists L = E(x) \quad \forall x \in \mathbb{P}_{r}: G(L) = x\]
This provides us with additional modeling capabilities that have proved successful in many papers in a variety of ways, \cite{infogan,vaegans,advae,began}. 

For us, the objective of inversion is important because it allows us to get the reconstruction of the original sample. We assert that if all the samples are perfectly reconstructed then their containing modes, respectively, can be safely said to have been captured. Usually, a modified form of direct modeling is applied to achieve this kind of inversion. The following losses can be added as additional objectives to achieve inversion, 
\[\min_{E}|x - G(E(x))| \quad \forall x \in \mathbb{P}_{r}\]
\[\min_{G}|z - E(G(z))| \quad \forall z \in \mathbb{P}_{z}\]
But this does not give a guarantee that a well-encoded mode will be generated. Thus, direct modeling of \(p(x|G(E(x)))\) can be added as an additional objective of the generator and vice-versa, this merges the two objectives as,
\[\min_{E,G} [|x - G(E(x)))| + |z - E(G(z))|], \quad \forall (x, z) \in (\mathbb{P}_{r}, \mathbb{P}_{z})\]
Optimizing this loss helps us achieve both our objectives, inversion and increases the probability of regeneration. A similar approach was proposed in CYCLEGANs \cite{cyclegan}. Two discriminators are used, one for each distribution, \(\mathbb{P}_{r}\) and \(\mathbb{P}_{z}\). They enforce the reconstruction objective by adding the mean-squared error to the vanilla adversarial objective. It should be noted that usually a uniform distribution is used as \(\mathbb{P}_{z}\). It can be observed that in this case there would be no mode loss in the Encoder output; since all possible points are equally probable, at the edge of a lost region, the gradients will always be present; and this is true for all dense or uni-modal distributions. Thus, the reconstruction loss is dropped for the Encoder, but we still include the term to preserve generality. Since \(\mathbb{P}_{z}\) is not a uniform distribution in the case of CYCLEGANs, but another set of natural images, cyclic consistency loss is added for Encoder too.

While direct modeling looks like a good solution, it interferes with the regular GAN \cite{gan} training. We explore this further in Appendix 1.2 in the supplementary material.

\subsection{BIGANs}
The BIGAN discriminator D discriminates not in the data space \(D(x)\) verses \(D(G(z))\), but jointly in data and latent space tuples \(D(x, E(x))\) versus \(D(G(z), z)\). The network is trained with the minimax objective,
\begin{multline*}
\min_{E,G}\max_{D} V(D,E,G) := \mathbb{E}_{x\sim p_{x}}[\log D(x,E(x))] + \\ \mathbb{E}_{z\sim p_{z}}[1 - D(G(z),z)]
\end{multline*}

It is shown that training using this objective, at the optimum, G and E naturally invert each other. An informal description of the objectives of the networks can be given as; the Encoder maps the samples \(x \in p(x)\) to \((x,E(x)) \text{ in } \mathbb{P}_{e}\), and the Generator maps the samples \(z \in p(z)\) to \((G(z),z) \in \mathbb{P}_{g}\). The two generated distributions, \(\mathbb{P}_{g}\) and \(\mathbb{P}_{e}\), are brought together using a classification or adversarial objective, assigning 1 to \(p_{g} \in \mathbb{P}_{g}\) and 0 to \(p_{e} \in \mathbb{P}_{e}\), the Generative networks, $G$ and $E$ are trained on the inverse objective.
It can be seen that, wherever the distributions intersect, inversion happens at the point. This is because at the intersection, \(z=E(x)\) and \(x=G(z)\), thus, \(x = G(E(x))\) and \(z = E(G(z))\). Since the BIGAN objective brings the two distributions together in the X-Z space, it naturally causes inversion. But, to achieve this objective \(\mathbb{P}_{e}\) will have to match \(\mathbb{P}_{z}\) and \(\mathbb{P}_{g}\) will have to match \(\mathbb{P}_{r}\). Thus, as a side-effect, the original GAN objective is also achieved.
But the Generator is still not free from mode loss; even if a mode is correctly encoded, the generator may lose it when the local gradients vanish. This is because the Generator improves with the objective as,
\[\min_{G} V(D,G) := \mathbb{E}_{z\sim p_{z}}[1 - D(G(z),z)]\] which does not apply a penalty for losing a mode. This is visually explained in section \ref{exp:2}

\subsection{Optimal Transport based GANs}
It is shown that optimizing the vanilla objective of the GAN is like minimizing Jensen-Shannon divergence between \(\mathbb{P}_{r}\) and \(\mathbb{P}_{g}\). Originally GANs were notorious for being difficult to train and required a balance to be maintained between the Generator and the Discriminator. WGANs \cite{wgan} proposed to change the objective from matching the probability distributions, \(\mathbb{P}_{r}\) and \(\mathbb{P}_{g}\), to a problem of optimal transport. The Discriminator is transformed to a critic which outputs an unbounded value, this is used to get the Earth Mover's distance between the real and generated distributions. This not only stabilized the training but the critic output correlates well with the sample quality. The objective function for WGANs is obtained by using Kantorovich-Rubinstein duality as,
\[\min_{G}\max_{D} V(D,G) = \mathbb{E}_{x\sim p_{x}}[D(x)] - \mathbb{E}_{z\sim p_{z}}[D(G(z))]\]
where D is the set of 1-Lipschitz functions. This was originally achieved via weight-clipping. Since WGANs involved directly modeling the gradients that are used by the generator to optimize, they allow for better control. An alternative to weight-clipping was proposed in Improved Training of WGANs \cite{iwgan}, where a gradient penalty (GP) is imposed on the Discriminator to maintain good gradients between real and fake data distributions.
\begin{multline*}
\mathcal{L}_{D} = \mathbb{E}_{x \sim \mathbb{P}_{g}}[D(x)] - \mathbb{E}_{z \sim \mathbb{P}_{z}}[D(G(z))] + \\ \lambda\mathbb{E}_{\tilde{x} \sim \mathbb{P}_{\tilde{x}}}[(\|\nabla_{\tilde{x}}D(\tilde{x}) - 1\|)]    
\end{multline*}
This was imposed at points sampled uniformly between the data points and the generations, \(\tilde{x}\). This works well if the distributions are well separated as then the mean direction of the gradient penalty is almost same for all points. However, once they overlap this is no longer true, and imposing a gradient between random unrelated points may not be beneficial.

\section{Our Method}

\subsection{Logit loss (LOGAN)}
We base our method on optimal transport as presented in WGANs \cite{wgan}. With optimal transport, we travel the discriminator values itself to optimize the generator.  This gives additional control over the gradients to allow for robust optimization. But we found that following the Wasserstein estimate and replacing the classifier with a critic can have limitations. To compensate for this while we can follow the gradients of the classifier based critic directly, there is a problem; With a sigmoid based discriminator, regions of saturation develop where it is easily able to classify the samples. This causes the local gradients for the generator to vanish, leading to slow and inefficient training. WGANs do not face this problem because the critic output is unbounded, thus there is no saturation. We propose a simple modification of the existing method; while we train the sigmoid based discriminator using the same objective,
\begin{multline*}
\mathcal{L}_{D} = \mathbb{E}_{\mathnormal{x \sim p_{X}}} [\log D(x, E(x))] + \\ \mathbb{E}_{\mathnormal{z \sim p_{Z}}} [\log (1 - D(G(z), z))]
\end{multline*}
 
We don’t follow its gradients directly for the generative network(s). Instead we use the logits of the discriminator outputs, thus LOGAN.
\begin{multline*}
\mathcal{L}_{E, G} = \mathbb{E}_{\mathnormal{x \sim p_{X}}} logit(D(x, E(x))) - \\ \mathbb{E}_{\mathnormal{z \sim p_{Z}}} logit(D(G(z), z))
\end{multline*}
Since the \(logit\) function is inverse of \(sigmoid\), it cancels the saturation that is induced in the classification based discriminator, and we get to keep our classifier.
Further it is noticeable that all the discriminator has to do is to assign \(1\) to \((x \in \mathbb{P}_{r})\) and \(0\) to \(G(z); z \in \mathbb{P}_{z}\). This is also achievable with the loss function,
\[ \mathcal{L}_{D} = D(G(z))^2 + (1 - D(x))^2 \]
\begin{itemize}
    \item This limits the maximum gradients from the \(\mathcal{L}_{D}\) to 1
    \item The gradients decay to 0 near the target value. This also regularizes the effect of the discriminator that can prevent an explosion in logit value, if any.
    \item This loss gives the same results at convergence.
\end{itemize}
While this looks similar to the loss function proposed in the LSGANs \cite{lsgan}, it is applied to the sigmoid outputs of the discriminator. The fact that the there is an upper bound on the gradients can be desirable in certain cases. We discuss an example in Appendix 1.2 in supplementary material.
We refer to the training done with formerly described discriminator objective as \(lol1\) and the latter as \(lol2\). We apply these losses combined with the gradient penalty \cite{iwgan}, to toy and natural image datasets. We show that this is not only an alternative to Wasserstein estimate but also provides better training with higher quality results. It was also observed that unlike WGANs, it is not required to update the \(D\) multiple times per \(G\) update in LOGANs.

\subsection{Pair-wise Gradient Penalty}

\begin{figure}[ht]
\centering
\begin{subfigure}{0.45\textwidth}
\centering
\includegraphics[width=\linewidth]{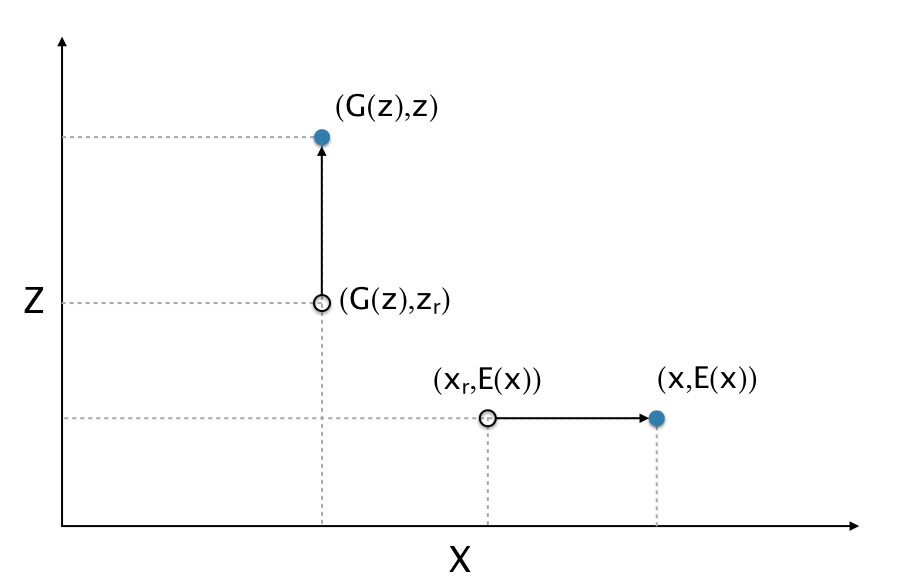} 
\caption{Application of pair-wise gradient penalty}
\label{fig:pwgp_a}
\end{subfigure}
\begin{subfigure}{0.45\textwidth}
\centering
\includegraphics[width=\linewidth]{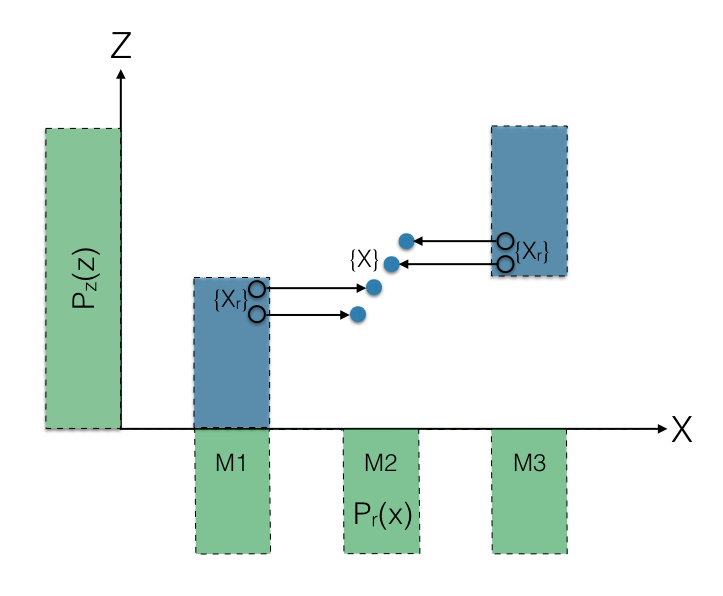}
\caption{A lost mode recovered using pair-wise gradient penalty}
\label{fig:pwgp_b}
\end{subfigure}

\caption{Illustrations of the application and the effect of pair-wise gradient penalty}
\label{fig:pw_grad_pen}
\end{figure}

As we stated in section 2.1, our objective involves encouraging accurate reconstructions. To achieve Encoder-Decoder inversion, we use a BIGAN with \(logit\) loss as our base model. But instead of applying a secondary objective such as mean-squared error for reconstruction, we directly model the gradients of the Discriminator to flow from the reconstructions to the data-points. Figure \ref{fig:pwgp_a} illustrates the application of pair-wise gradient penalty to 1-D X and Z. If a mode is lost, its reconstructions will lie in a nearby major mode, if the gradients are enforced between the reconstructions and the real data points, the lost mode can be recovered, figure \ref{fig:pwgp_b} illustrates this for a toy tri-modal dataset, where the middle mode is lost, but then the reconstructions are pulled from the nearby modes to it.

To present this more formally we extend from the argument presented in Appendix 1 of Improved training of WGANs, \cite{iwgan}. The optimal coupling between \(\mathbb{P}_{r}\) and \(\mathbb{P}_{g}\) is defined as,
\[
W(\mathbb{P}_{r}, \mathbb{P}_{g}) = \inf_{\pi \in \Pi(\mathbb{P}_{r}, \mathbb{P}_{g})} \mathbb{E}_{(x,y)\sim \pi} [\|x - y\|] 
\]
Where \(\Pi (\mathbb{P}_{r} , \mathbb{P}_{g})\) is the set of joint distributions \(\pi(x, y)\) whose marginals are \(\mathbb{P}_{r}\) and \(\mathbb{P}_{g}\), respectively.

Let \(\mathbb{P}_{\hat{r}}\) be the distribution of the reconstructions generated as \(\hat{x} = G(E(x)), \forall x \in \mathbb{P}_{r}\). Since the optimal encoder and decoder BIGANs naturally invert each other, under optimal conditions,
\[\mathbb{E}_{x\sim \mathbb{P}_{r}} [\|x - G(E(x))\|] \rightarrow 0\]

Thus, the reconstructions can serve as the optimal couplings for \(x \in \mathbb{P}_{r}\). We can rewrite the equation as,
\[
W(\mathbb{P}_{r}, \mathbb{P}_{\hat{r}}) = \inf_{\pi \in \Pi(\mathbb{P}_{r}, \mathbb{P}_{\hat{r}})} \mathbb{E}_{(x,y)\sim \pi} [\|x - y\|] 
\]

Instead of applying a gradient penalty between random pairs \((x, G(z))\), we place a gradient penalty between optimal couplings \((x, G(E(x))), x \in \mathbb{P}_{r}\) and \((E(G(z)), z), x \in \mathbb{P}_{z}\). Since we are using BIGANs, we work in a joint space of X and Z, the gradient penalty equations are modified as,

\[\epsilon \in U[0,1], x \sim \mathbb{P}_{r}\]
\[\hat{x} = G(E(x))\]
\[\tilde{x} = \epsilon x + (1-\epsilon)\hat{x}\]
\[x_{unit} = \frac{x - \hat{x}}{\|x - \hat{x}\|}\]
\[\mathcal{L}_{GP_{x}} = \mathbb{E}_{x \sim \mathbb{P}_{r}} [\|\nabla_{\tilde{x}} D(\tilde{x}, E(x)) - x_{unit}\|]\]

Similar gradient penalty can be applied for the other distribution.

\[\epsilon \in U[0,1], z \sim \mathbb{P}_{z}\]
\[\hat{z} = E(G(z))\]
\[\tilde{z} = \epsilon z + (1-\epsilon)\hat{z}\]
\[z_{unit} = \frac{z - \hat{z}}{\|z - \hat{z}\|}\]
\[\mathcal{L}_{GP_{z}} = \mathbb{E}_{z \sim \mathbb{P}_{z}} [\|-\nabla_{\tilde{z}} D(G(z), \tilde{z}) - z_{unit}\|]\]

Therefore, the total loss for the discriminator is,

\[\mathcal{L}_{D} = \mathcal{L}_{adv} + \lambda( \mathcal{L}_{GP_{x}} + \mathcal{L}_{GP_{z}})\]

The pair-wise gradient penalty is in both direction and magnitude. If \(\mathbb{P}_{z}\) is a uniform distribution, gradient penalty \(\mathcal{L}_{GP_{z}}\), is not required. It should be noted that since we train the encoder to minimize \(\mathbb{E}_{\mathnormal{x \sim p_{X}}} logit(D(x, E(x)))\), it travels down the discriminator terrain, so we need to put a gradient penalty in the opposite direction to the gradients, so we place a negative sign in the equation for \(\mathcal{L}_{GP_{z}}\). This establishes a gradient between \((x, z) \in (\mathbb{P}_{r}, \mathbb{P}_{z}) \) and \((\hat{x}, \hat{z}) \in (\mathbb{P}_{\hat{r}}, \mathbb{P}_{\hat{z}})\) that guides the reconstructions to their optimal coupling.

\begin{figure*}[ht]
 
\begin{subfigure}{0.24\textwidth}
\centering
\includegraphics[width=\linewidth]{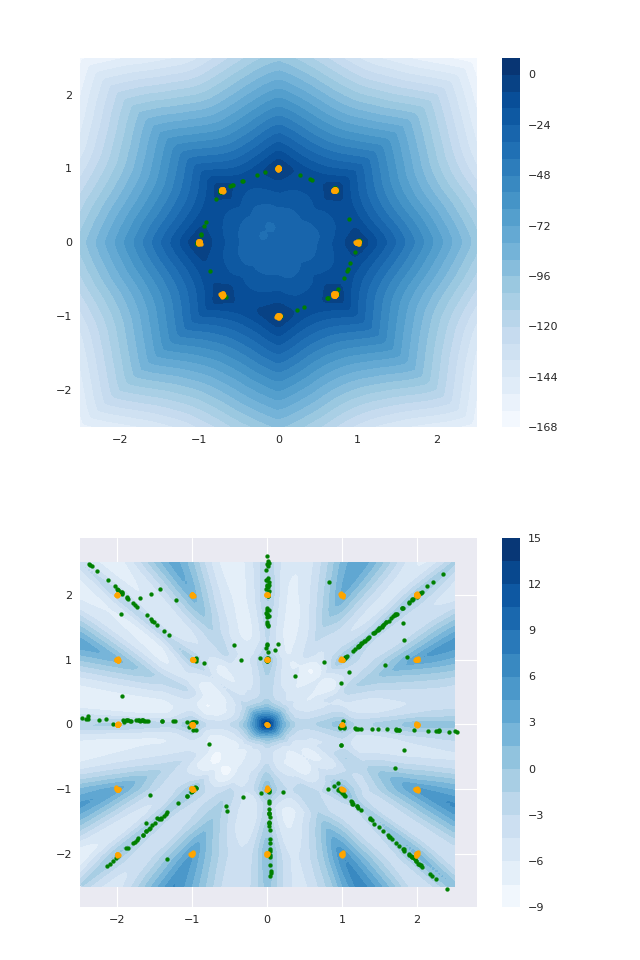} 
\caption{vanilla}
\label{fig:2dv}
\end{subfigure}
\begin{subfigure}{0.24\textwidth}
\centering
\includegraphics[width=\linewidth]{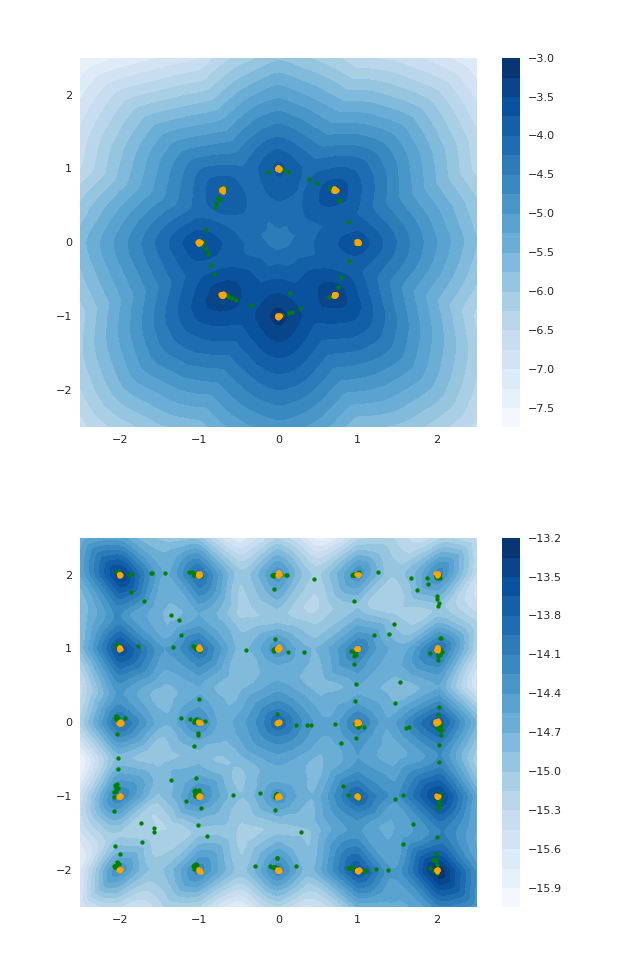}
\caption{wasserstien; GP}
\label{fig:2dw}
\end{subfigure}
\begin{subfigure}{0.24\textwidth}
\centering
\includegraphics[width=\linewidth]{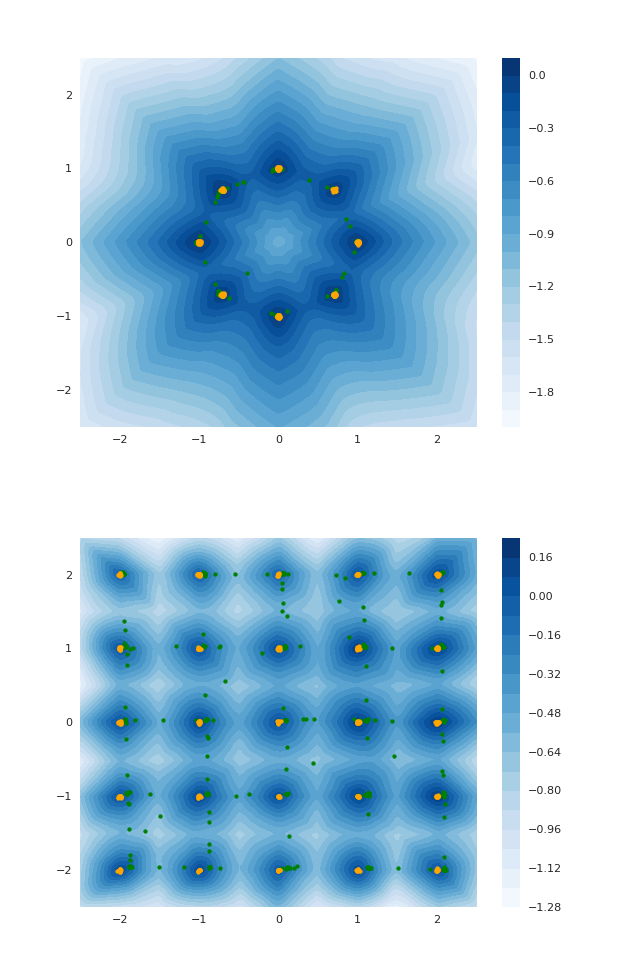}
\caption{lol1; GP}
\label{fig:2dlol1}
\end{subfigure}
\begin{subfigure}{0.24\textwidth}
\centering
\includegraphics[width=\linewidth]{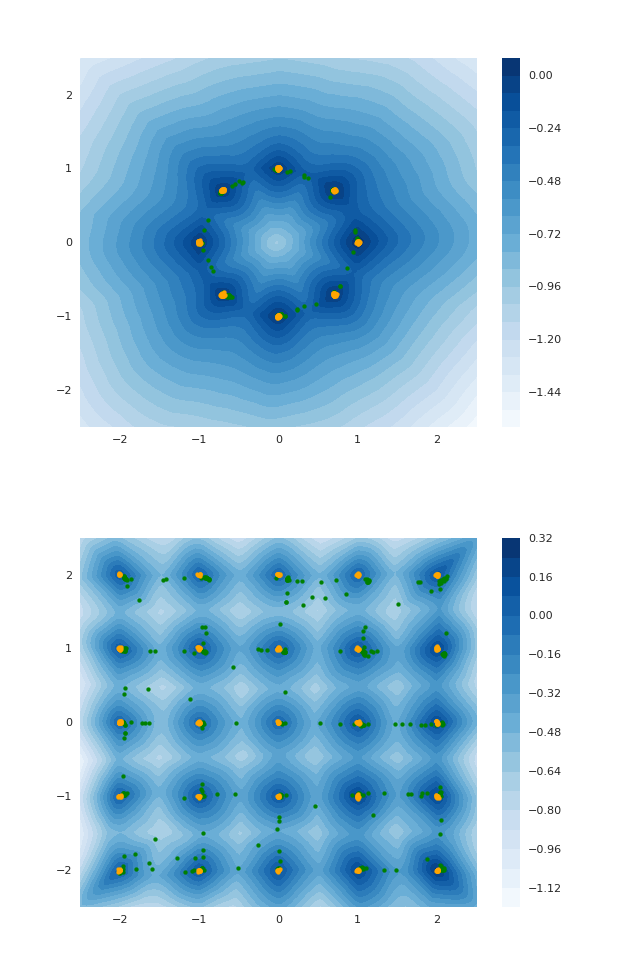}
\caption{lol2; GP}
\label{fig:2dlol2}
\end{subfigure}
 
\caption{Discriminator contour patterns generated while training a model on 2D toy datasets. It can be seen that using logit loss results in much better and uniform gradient contours. The contours are drawn using the pre-activation outputs from the Discriminator}
\label{fig:2dplots}
\end{figure*}

\subsection{Additional benefits}

\begin{figure*}[ht] 
 
\begin{subfigure}{0.24\textwidth}
\centering
\includegraphics[width=\linewidth]{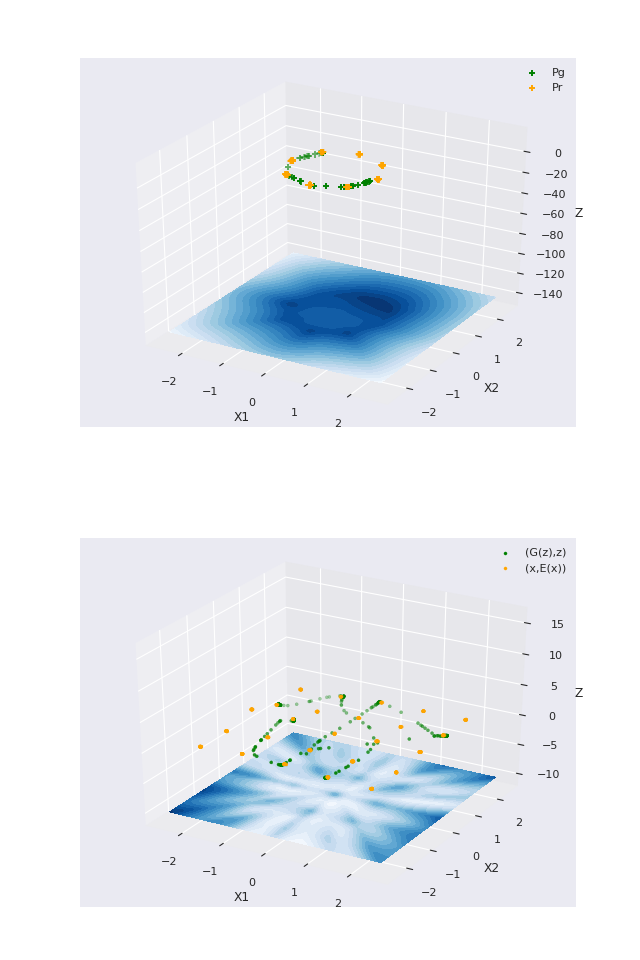} 
\caption{vanilla}
\label{fig:3dv}
\end{subfigure}
\begin{subfigure}{0.24\textwidth}
\centering
\includegraphics[width=\linewidth]{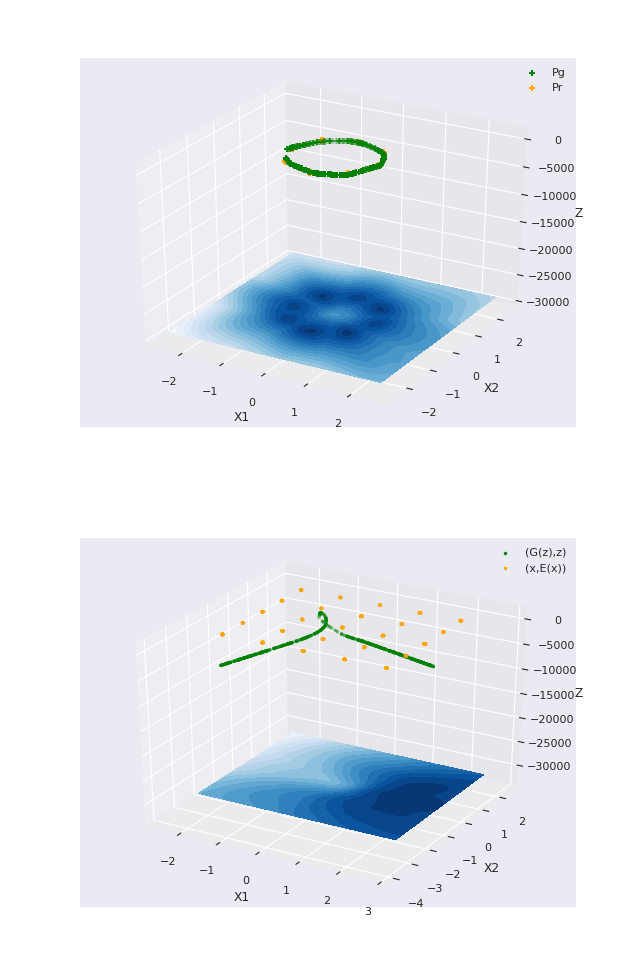}
\caption{wasserstien; pair-wise GP}
\label{fig:3dw}
\end{subfigure}
\begin{subfigure}{0.24\textwidth}
\centering
\includegraphics[width=\linewidth]{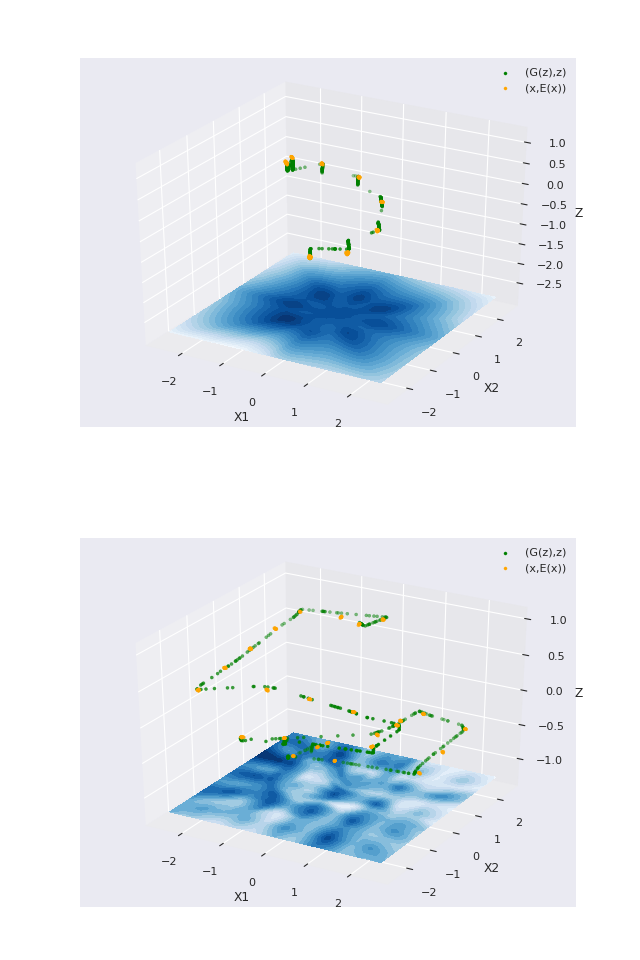}
\caption{lol1; pair-wise GP}
\label{fig:3dlol1}
\end{subfigure}
\begin{subfigure}{0.24\textwidth}
\centering
\includegraphics[width=\linewidth]{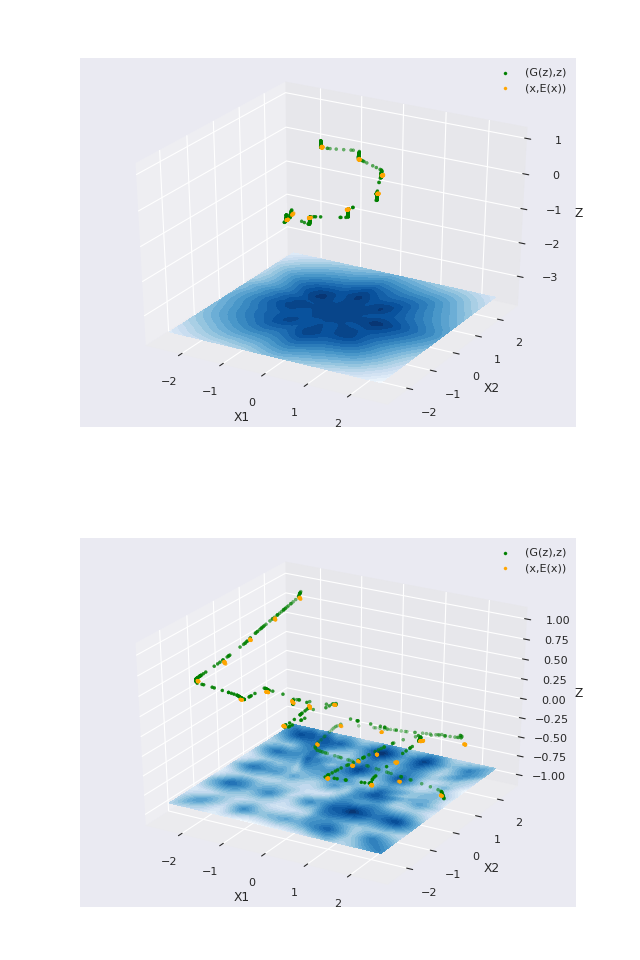}
\caption{lol2; pair-wise GP}
\label{fig:3dlol2}
\end{subfigure}
 
\caption{BIGAN trained on toy datasets. It can be seen that while the model faces heavy model loss with only the vanilla objective, it trains quite well with the application of logit loss with pair-wise gradient penalty}
\label{fig:3dplots}
\end{figure*}

In regular GAN training, once a mode collapse happens, it cannot recover, because the Generator experiences same gradients for each of its generated samples. Thus, the point distribution \(\mathbb{P}_{g}(z)\) wanders in the X space to settle to a local optima.
But in the case of BIGANs, if a dense distribution is used for \(p(z)\), like uniform, then as stated in section 2.1, the Encoder will not collapse; or will easily recover from a mode collapse. Then even if we start off with a collapsed Generator, it can be reversed. This is because if the Encoder is able to well encode, ie. \(X \mapsto E(x)\), then the gradients experienced by the Generator will be different for each of its generated samples, \((G(z),z)\) to \((x,E(x))\). Thus the collapsed \(p(g)\) is stretched to the other modes. This breaks the collapse to some extent, and the Generator with vanilla objective manages to catch the nearby still reachable modes.
This can be visualized using figure \ref{fig:3dplots}, for a collapsed Generator, the \(green\) "string" \((G(z),z)\), would be a perfectly straight vertical line; but the \(yellow\) "dots" \((x,E(x))\), would remain at the same places as the Encoder will be well-encoded. With training the string would bend by moving samples in the X-plane and reach the nearby modes.
With the application of pair-wise gradient penalty, gradients would be explicitly created from \((G(E(x)), E(x))\) to \((x, E(x))\). This would further assist the "string" to bend even more guiding the reconstructions to the actual samples. Thus, the modified gradients help the Generator to reach to all the modes well-Encoded by the Encoder.
In cases when \(\mathbb{P}_{z}\) is not a dense distribution, the gradient penalty is applied in both directions, X and Z, as described in section 3.2.

It should be noted that using the \(logit(D)\) to implement the gradient penalty has a benefit that we do not have to scale the data space to compensate for the limits on the D value, i.e., if \(x \in [-1,1]^N\) then, a constant gradient of 1 from point \([-1]^N\) to \([1]^N\) with \(D(x) \in (0,1)\) cannot exist.

An optimal discriminator tries to \(logit(D(x)) \rightarrow \infty\) and \(logit(D(G(z))) \rightarrow -\infty\), and since the gradients decrease as the value approaches the objective, it offers a kind of regularization causing the \(logit\) values of the real samples and fake samples to lie symmetrically about the origin. But, this is more of a cosmetic benefit.

In Experiment 4.2 we show that while the network fails to train with the Wasserstein estimate, LOGAN works well. Sometimes it can be desirable to have a classifier as a discriminator over a critic, but this will be more of a task and preference dependent benefit. But if an unsaturated critic value is required, the \(logits\) of the discriminator D can be used.

\section{Experiments}
We provide details regarding the network architectures and the parameter settings that we have used for the experiments in the supplementary material. In this section, we first examine the effectiveness of LOGAN using toy datasets that clearly illustrate the improved performance. We then examine the effect of gradient penalty. Next, we present results on reconstruction and generation tasks. We further study the effect of an adverse surveillance dataset with a large single central mode. Finally, we evaluate the qualitative improvements in results.

\subsection{Effectiveness of LOGAN} \label{exp:1}

We test our training objective on 2 toy datasets with scattered modes. We use a 4 layered MLP with LeakyRELU activations as our model. Z is sampled from a uniform distribution \([U(-1,1)]^2\). We compare our results against the same model trained with the vanilla objective \((v)\), and the Wasserstein objective \((w)\). A gradient penalty, as described in \cite{iwgan}, is applied as an additional objective for Wasserstein and logit loss. Figure \ref{fig:2dplots} shows the trained discriminator value contours and the trained generator distributions. It can be seen that the discriminator trained with logit loss attains more uniform and cleaner patterns.

\begin{figure}
    \centering
    \includegraphics[width=0.43\textwidth]{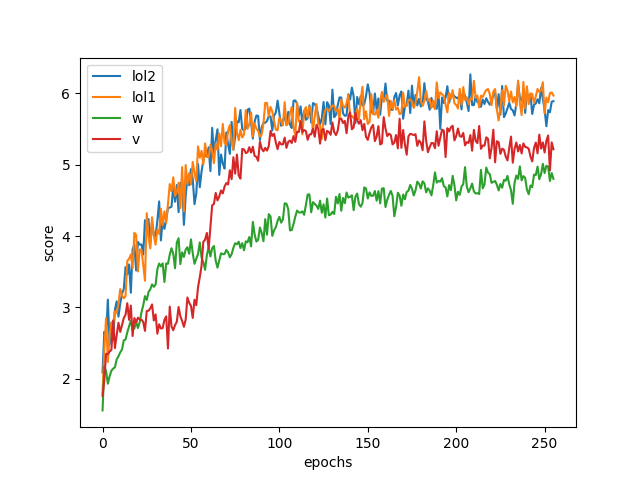}
    \caption{Inception scores attained by different losses on the BIGAN model for the CIFAR 10 dataset}
    \label{fig:bi_inc}
\end{figure}

To measure the model performance on natural images we train a DCGAN network with the three losses. \(z\) is sampled from a uniform distribution \([U(-1,1)]^{128}\). 
We list the inception scores calculated for different losses in Table \ref{table:dc_inc_scores}, since DCGAN has a similar network as ours, we list the scores reported by it for comparison. The details about the model architecture and training method are given in Appendix 1.1 in Supplementary material.

\begin{table}
\begin{center}
\begin{tabular}{ c c }
\hline
Method & Score \\
\hline
DCGAN & $6.16 \pm .07$ \\
LOGAN lol1 & $6.52 \pm .15$ \\
LOGAN lol2 & $6.55 \pm .19$
\end{tabular}
\caption{Inception scores attained by our models using DCGAN as the base architecture}
\label{table:dc_inc_scores}
\end{center}
\end{table}

\subsection{Effectiveness of Pair-wise gradient penalty} \label{exp:2}

To illustrate the effectiveness of pair-wise gradient penalty visually we use 2-D toy data sets and a 1-D z from a uniform distribution. The distributions are then drawn in a 3D space. The \((x,E(x))\) are plotted in "yellow" and \((G(z),z)\) in "green", figure \ref{fig:3dplots}. It can be seen how the Encoder tries to move the X points in the Z space to match \(p(z)\), here uniform. The Generator tries to map the Z in X space to match the \(p(x)\) distribution. This in effect causes the point cloud of the 2 class to come together, and under ideal convergence, the two should exactly match resulting in perfect inversion. Due to the limited and clustered nature of the modes of our toy dataset, the yellow point cloud appears as "dots"; And because \(p(z)\) is uniform and continuous, the green point cloud is like a "string". The bottom plane is the contour plot of \(D(x, E(x))\), it can be thought of as a 2-D projection of 3-D \(D(x,z)\). We are concerned with the X plane only because \(p(x)\) is the only distribution that needs examination for mode loss. We use same toy distributions from the previous section to train an encoder, a decoder and a discriminator with the same losses. We observed that the network trained with the vanilla loss consistently lost a few modes. The training was unstable with the Wasserstein estimate; It converged only at the very last steps of the training for the circular dataset and does not converge for the square. With the application of the pairwise gradient penalty, coupled with logit loss, it can be seen that almost all of the modes are captured.

\subsection{Quantitative evaluation of reconstruction and generation tasks}
Measuring similarity between the data and the reconstructions of natural images has been a difficult task. It has been shown in the past that the mean squared error is not a good measure of similarity of images \cite{vaegans}. We use Structural-Similarity (SSIM) \cite{ssim} index to judge the similarity of the real and fake samples. Specifically, we compare the moving average of the values from SSIM in order to not be sensitive to a particular mode being captured well. It was noticed that our methods edge the vanilla method by a small margin, Table \ref{table:bi_ssim}.

\begin{table}
\begin{center}
\begin{tabular}{ c c }
\hline
Method & SSIM \\
\hline
Vanilla & $0.223 \pm .08$ \\
LOGAN lol1  & $0.238 \pm .09$ \\
LOGAN lol2  & $0.230 \pm .09$
\end{tabular}
\caption{Best moving average values achieved on BIGAN using different losses}
\label{table:bi_ssim}
\end{center}
\end{table}

To judge the improvement in generation quality, if any, we compare the inception scores of the generations from the BIGAN with vanilla objective and LOGAN in Fig 5. We observe that the curves from the proposed method consistently are better than the other models across epochs. Table \ref{table:bi_inc_scores} lists inception scores from 50k samples of the CIFAR 10 dataset from different models. We find that there is a noticeable difference in the results between models using and not using the gradient penalty. We believe that this is explainable by the fact that since \(\mathbb{P}_{z}\) is continuous and the Generator is just a function that maps it to the X-space, thus \(\mathbb{P}_{g}\) is also continuous, leading to the generation of unreal looking samples from between the modes. The probability of generation of such a sample increases with the number of modes that are captured, leading to a lower score.

\begin{table}
\begin{center}
\begin{tabular}{ c c }
\hline
Method & Score \\
\hline
Vanilla & $5.31 \pm .07$ \\
Wasserstein & $4.8 \pm .15$ \\
LOGAN lol1  & $6.11 \pm .05$ \\
LOGAN lol2  & $6.23 \pm .03$
\end{tabular}
\caption{Inception scores attained by our models with BIGAN as the base architecture}
\label{table:bi_inc_scores}
\end{center}
\end{table}

\subsection{Single central mean dataset}

\begin{figure*}[ht]
 
\begin{subfigure}{\textwidth}
\centering
\includegraphics[width=\linewidth]{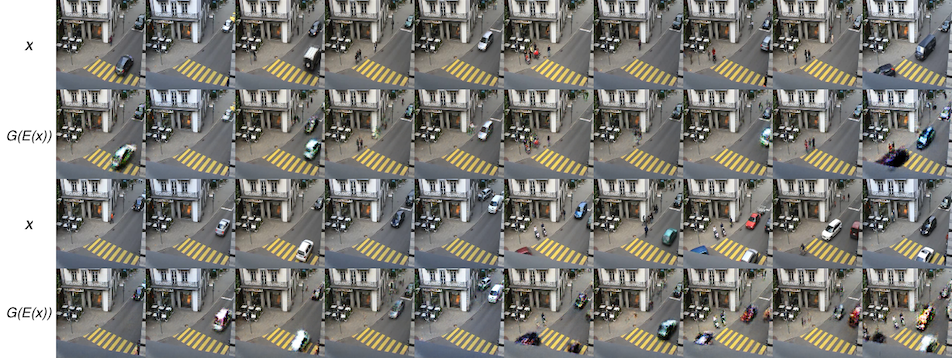}
\caption{Sample reconstructions of the input images from the traffic dataset}
\label{fig:traffic_recon}
\end{subfigure}
\begin{subfigure}{\textwidth}
\centering
\includegraphics[width=\linewidth]{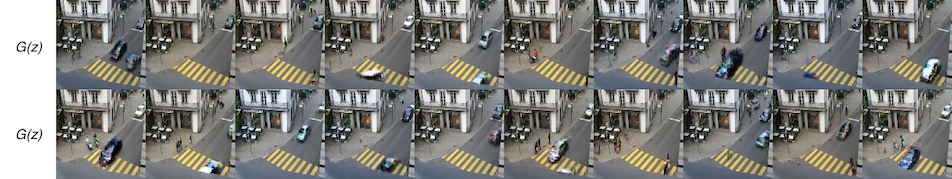}
\caption{Sample generations from our model trained on the traffic dataset}
\label{fig:traffic_gen}
\end{subfigure}
 
\caption{}
\label{fig:traffic}
\end{figure*}

\begin{figure}
    \centering
    \includegraphics[width=\linewidth]{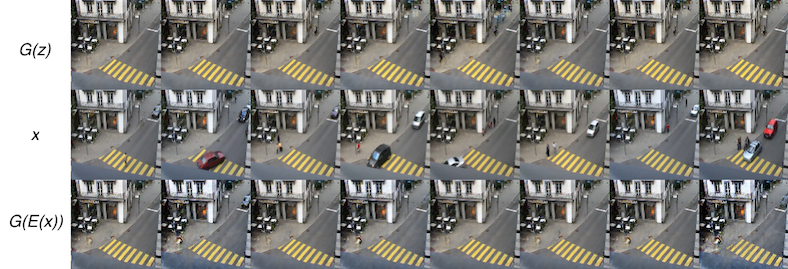}
    \caption{Generations and reconstructions from a BIGAN trained with the vanilla objective. It can be easily seen that the generator has collapsed to the mean}
    \label{fig:v_traffic}
\end{figure}

We create a dataset by taking frames from a surveillance footage of traffic junction at 5 fps. Originally this dataset \cite{traffic} was meant for abnormal activity detection. We observed that the frames mostly featured very limited activity. Some activity was replicated in multiple frames and some existed for only a few. This was because the activity in the video slowed down sometimes, leading to multiple frames of the same kind, and when the activity went smoothly, the opposite was true. The samples from the image distribution lie very close to one another; very few extremely strong modes and some really weak ones. This dataset was chosen to test the network for mode collapse as the clustered nature of the dataset heavily encourages it. We train the BIGAN model with the vanilla loss and our proposed method. We observed that the vanilla implementation consistently captured a few modes in the beginning but very quickly collapsed to a single mode, Figure \ref{fig:v_traffic}. But the same model trains to capture a wide variety of modes with our proposals. Figure \ref{fig:traffic}, shows samples at convergence. While the results here are not perfect, clearly they show that the model with our method indeed tries to capture all the modes.

\subsection{Qualitative Assessment} \label{exp:3}

As the inception score is a soft metric we encourage the qualitative assessment of the results too. It should be noted that our main contribution is not the improvement of the quality or the stability of the process, but to try to capture various minor modes that are lost in the vanilla process. Thus, we would want to look for hints of diversity in the generations and accuracy in the reconstructions produced by the network. Since we could not find an accurate enough metric to capture this, we rely on manual assessment.

\begin{figure}
    \centering
    \includegraphics[width=\linewidth]{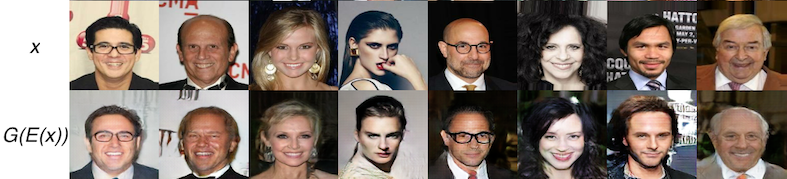}
    \caption{Reconstructions of samples from a BIGAN based on our loss, trained on CelebA dataset.}
    \label{fig:celeb_recon}
\end{figure}

We train a DCGAN network on 128x128 CelebA \cite{celeba} to show the quality and diversity of our samples. We try to keep the architecture of the network simple, we do not use any skip layers or residual blocks, this is to keep from any bias that may reflect in the quality of our results. Figure \ref{fig:celeb_recon} shows the reconstructions generated from the CelebA dataset, it can be seen that the model picks up minor features like optical glasses, sunglasses, background with patterns, shadows behind the faces, hats, earrings (not distinctly visible, mostly white scribbles), clothes and even a defect that was present on the top side of certain images. 

We also show the generations from the interpolations of the latent representations of the images to show that the model learns meaningful features of the faces, Figure \ref{fig:celeb_inter}. It can be noticed that the model not only learns the high-level features like the pose, skin tone, gender, ethnicity, etc. but also learns to smoothly transit between the minute features like caps, hats, sunglasses, optical glasses, etc. Additional results for such examples are provided in supplementary material. It should be noted that almost all of the interpolations are clear, none are blurry, this indicates a better convergence of the network. A hint of this can be seen in figure \ref{fig:celeb_inter}, where the glasses of the girl are removed in an instant and not faded away, indicating that the model has learned a better separation between the two modes, with and without glasses.

\begin{figure}
    \centering
    \includegraphics[width=\linewidth]{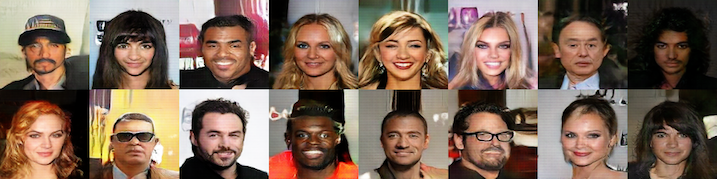}
    \caption{Generations from a BIGAN based on our loss, trained on CelebA dataset. It can be observed from the diversity of the samples that the model generates even the minor learned features.}
    \label{fig:celeb_gen}
\end{figure}

\begin{figure}
    \centering
    \includegraphics[width=\linewidth]{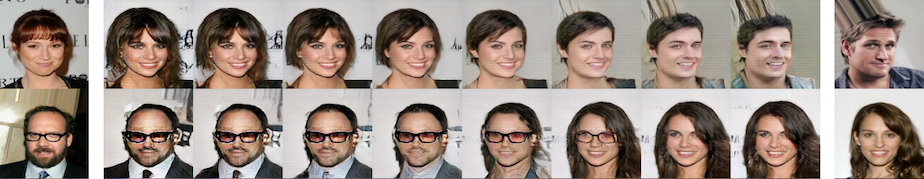}
    \caption{Generations from interpolations of the latent representations of the images. Note the smooth transition between female and male, but the discontinuous transition between having eyeglasses and not having eyeglasses.}
    \label{fig:celeb_inter}
\end{figure}

\section{Conclusion}

In our proposal, we introduce methods that directly aim to capture the full data distribution. We show that these are able to capture the weaker modes while being stable to train. These modifications are possible based on our loss functions. The proposed method LOGAN can also be viewed as an alternative to WGAN to model the problem of optimal transport. We show via toy datasets and CIFAR inception score that the proposed method is indeed superior both quantitatively and qualitatively. We also additionally incorporated {\it pair-wise} gradient penalty for BIGANs trained with optimal transport, that helps the network to reach weaker modes resulting in diverse and visually pleasing images. To conclude, we show that with modified loss functions we can capture the data distribution without losing modes.

\bibliographystyle{aaai}
\bibliography{references}

\section{Appendix}

\subsection{Network Architecture and Training details}

We try to keep our architectures as primitive as possible to keep from any biases that may appear in the results from it. All training procedures use an Adam Optimizer \((\alpha=0.0001; \beta_{1}=0.5; \beta_{2}=0.9)\).
For toy datasets we use:
\begin{itemize}
    \item A 4-layered MLP with LeakyRELU activations as our network for both Generator and the Discriminator.
    \item No batch normalization in either Generator or Discriminator, we observed it proved unfruitful in all cases.
    \item For Experiment 4.1 we use 2-D X and a 2-D uniform distribution \([U(-1,1)]^2\). For the 3-D plots in Experiment 4.2, we use a 1-D Z from \([U(-1,1)]\).
    \item The network is trained for 40000 steps
    \item For gradient penalty, \(\lambda=.1\) for all toy datasets.
    \item We found that in case of toy datasets, initializing the weights using a random uniform distribution as described in \cite{uni_init} improved the stability significantly.
\end{itemize}
For image architectures we use:
\begin{itemize}
    \item A basic linear layer followed by fractionally-strided convolutional networks, as proposed in DCGAN \cite{dcgan}, with LeakyRELU activations. Same architecture is used for the Generator in the BIGAN model, and its inverted form for the Encoder.
    \item We apply batch normalization in the Generator at every layer but the last. An inverse of the Generator architecture is used for the Encoder for the experiments requiring it.
    \item Z is sampled from a uniform 128-dim space \([U(-1,1)]^{128}\)
    \item We run the training for 256 epochs for each image dataset.
    \item \(\lambda=1\), for all image datasets.
\end{itemize}

We found that removing batch normalization from the Discriminator helped improve all our loss cases. For training, we do not apply any extra regularization like Historical averaging, mini-batch discrimination \cite{improved_gan}. The Discriminator and Generator are updated once per step, except in the case of Wasserstein where we update \(D\) five times per \(G\) iteration. For further details, please refer to our open source implementation \url{https://github.com/shashank879/logan}.

\subsection{Comparison with direct modeling}

\begin{figure}
    \centering
    \includegraphics[width=\linewidth]{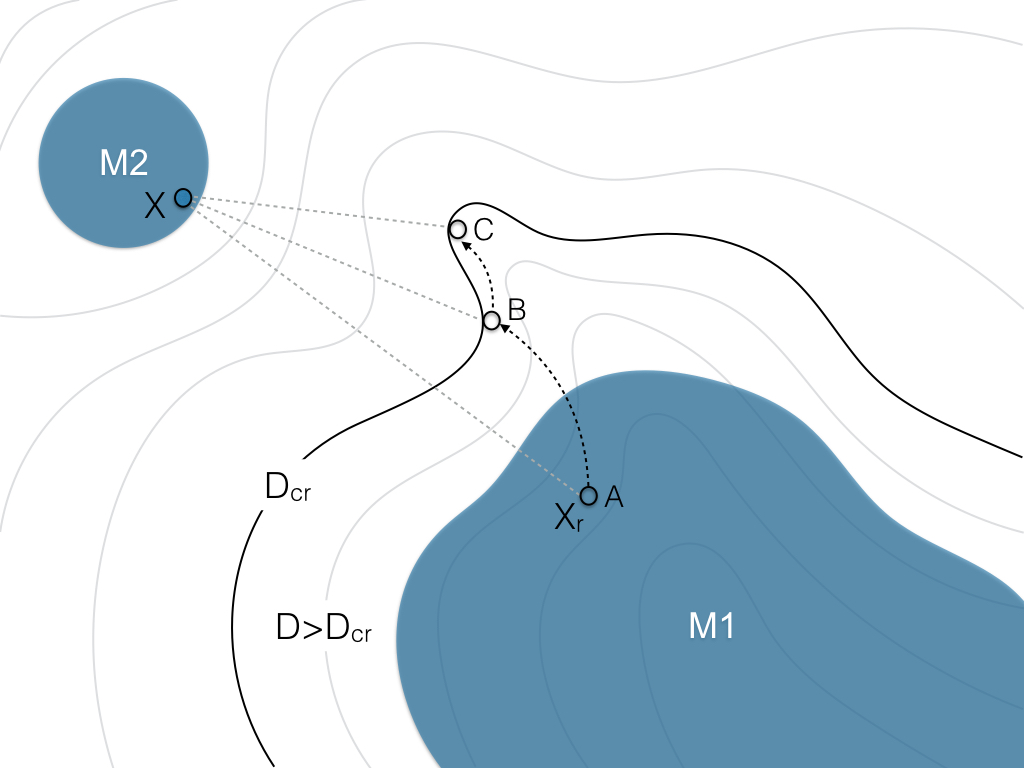}
    \caption{An illustration showing how the mse loss with bounded gradients cannot overcome unbounded adversarial gradients, leading to failure in mode capture}
    \label{fig:ae_mode_loss}
\end{figure}

We consider a model based on BIGAN, but mean-squared error between the data and its reconstructions is added as the modeling objective for the encoder and the decoder as:

\[\min_{G} [|x - G(E(x)))|], \quad \forall x \in \mathbb{P}_{r}\]
\[\min_{E} [|z - G(E(z)))|], \quad \forall z \in \mathbb{P}_{z}\]
While this looks like a good solution, it interferes with the regular GAN training. A balance has to be maintained between the losses, generally achieved through a fixed ratio between the losses. It is difficult to be sure whether a given ratio for the losses would be enough to pull the reconstruction to the actual data from the adversarial optima. \(\mathcal{L}_{adv}\) is unbounded but, the \(\mathcal{L}_{recons}\) is usually bounded as the dimensions have limits. Thus, if the region between \^{x} and x is rejected by the discriminator with a high enough confidence, i.e. \(D(G(z))\) is low for the region between them, the reconstruction will not be able to cross to the coupling mode. This is illustrated in Figure \ref{fig:ae_mode_loss}

If \(x \in [-1,1]^N\), then the maximum reconstruction loss, if measured as L1, is given as 2N, and provides the gradient with magnitude of \(2N\) everywhere. Then, for the reconstruction loss to overcome the adversarial loss,

\begin{align*}
|\lambda\nabla_{x}\mathcal{L}_{recons}| &\geq |\nabla_{x}\mathcal{L}_{adv}|\\
2N\lambda &\geq |\frac{-1}{D(G(z))}|\\
D(G(z)) &\geq \frac{1}{2N\lambda}\\
D_{crit} &= \frac{1}{2N\lambda}
\end{align*}
Let \(x \in \mathbb{P}_{r}\) and \(y \in \mathbb{P}_{\hat{r}}\), assume that at convergence the mode of x is not captured and y, the reconstruction of x, lies in a different mode. Since x and y though separated, lie near modes, thus the most probably the D values at \(x\) and \(y\) are,
\[D(x) > D_{crit} \quad D(y) > D_{crit}\]
Then if there does not exist a path, \(P(e): e \in [0,1]\), from x to y such that, \(P(0) = x\) and \(P(1) = y\), and,
\[D(P(e)) > D_{crit} \quad \forall e \in [0,1]\]
The mode cannot be captured by the proposed loss reconstruction loss. Our proposal of \(lol2\) gives a bounded adversarial loss which can be useful in such a case. But due to lack of time we could not fully explore the possibilities with their union. Training with default values was unstable though, it may be due to the improper balancing of the objectives.

To summarize, while the generator-discriminator interactions from the adversarial loss and the mean-squared error are individually predictable, we found that but the mixture of both is not and hard to control.

It can also be shown that our method does not suffer from a effect similar to the one that arises from bounded reconstruction loss. It is because our method does not try to overcome the adversarial objective but rather work with it to acheive the objective.
Given the same situation, since \(y\) is near a mode, \(D(y)\) will generally be high, still \(logit(D(x))\) can always be increased such that the gradient penalty, \(\min_{D}[\nabla_{x} D(\tilde{x}, E(x)) - x_{unit}]^2\), where \(x_{unit} = \frac{x-y}{\|x-y\|}\), is satisfied. And since it aligns with the adversarial objective, \(\max_{\mathcal{D}}[\log{D(x)}]\), they can be satisfied simultaneously.

\section{Additional samples}

\begin{figure}[!ht]
\centering
\begin{subfigure}{0.45\textwidth}
\centering
\includegraphics[width=\linewidth]{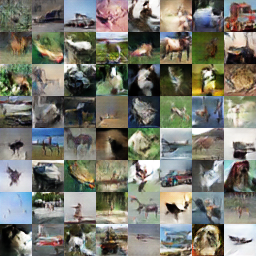} 
\caption{lol1}
\label{fig:dc_lol1}
\end{subfigure}
\begin{subfigure}{0.45\textwidth}
\centering
\includegraphics[width=\linewidth]{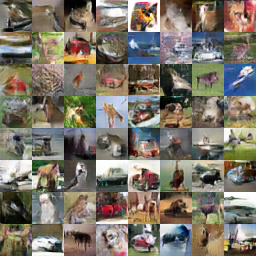}
\caption{lol2}
\label{fig:dc_lol2}
\end{subfigure}
 
\caption{Sample generations from a DCGAN-like network trained with our losses}
\label{fig:dc_samples}
\end{figure}

\begin{figure*}[!ht]
\centering
\begin{subfigure}{0.45\textwidth}
\centering
\includegraphics[width=\linewidth]{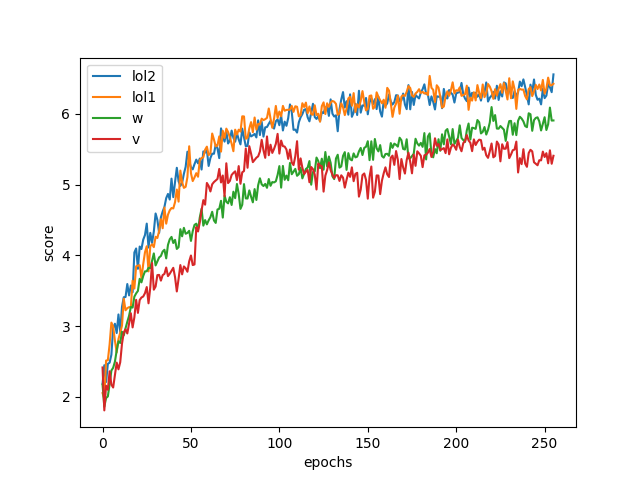}
\caption{Regular GAN}
\label{fig:dc_inc}
\end{subfigure}
\begin{subfigure}{0.45\textwidth}
\centering
\includegraphics[width=\linewidth]{figures/graphs/bi_inc.png}
\caption{BIGAN}
\label{fig:bi_inc}
\end{subfigure}
\caption{Cifar10 Inception scores attained by regular GAN and BIGANs during training with different losses}
\label{fig:inc_score}
\end{figure*}

\begin{figure*}[!ht]
\centering
\begin{subfigure}{0.45\textwidth}
\centering
\includegraphics[width=\linewidth]{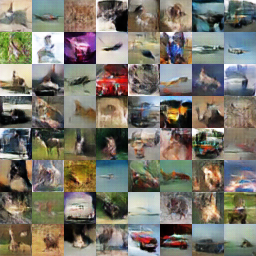} 
\caption{lol1}
\label{fig:bi_lol1_gens}
\end{subfigure}
\begin{subfigure}{0.45\textwidth}
\centering
\includegraphics[width=\linewidth]{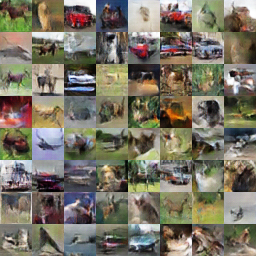}
\caption{lol2}
\label{fig:bi_lol2_gens}
\end{subfigure}
 
\caption{Sample generations from a deep-convolutional BIGAN network trained with our losses}
\label{fig:bi_samples}
\end{figure*}

\begin{figure*}
\centering
\begin{subfigure}{0.45\textwidth}
\centering
\includegraphics[width=\linewidth]{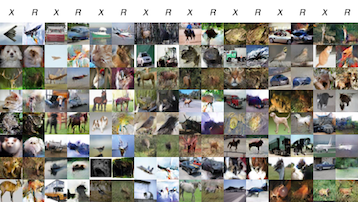} 
\caption{lol1}
\label{fig:bi_lol1_recons}
\end{subfigure}
\begin{subfigure}{0.45\textwidth}
\centering
\includegraphics[width=\linewidth]{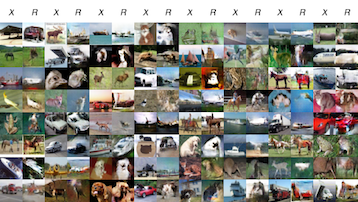}
\caption{lol2}
\label{fig:bi_lol2_recons}
\end{subfigure}
 
\caption{Sample reconstructions from a deep-convolutional BIGAN network trained with our losses}
\label{fig:bi_recons}
\end{figure*}

\begin{figure*}
    \centering
    \includegraphics[width=\linewidth]{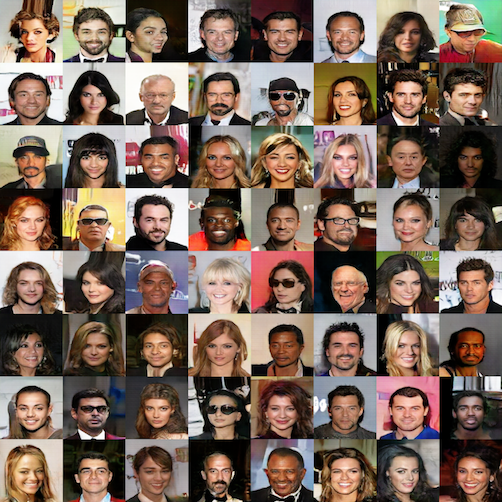}
    \caption{Sample generations from a deep-convolutional BIGAN trained on the CelebA dataset.}
    \label{fig:celeba_gens}
\end{figure*}

\begin{figure*}
    \centering
    \includegraphics[width=\linewidth]{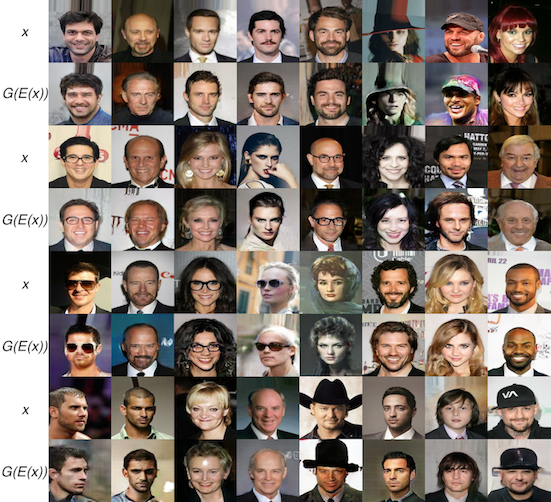}
    \caption{Sample reconstructions from a BIGAN trained on the CelebA dataset.}
    \label{fig:celeba_recons}
\end{figure*}

\begin{figure*}
    \centering
    \includegraphics[width=\linewidth]{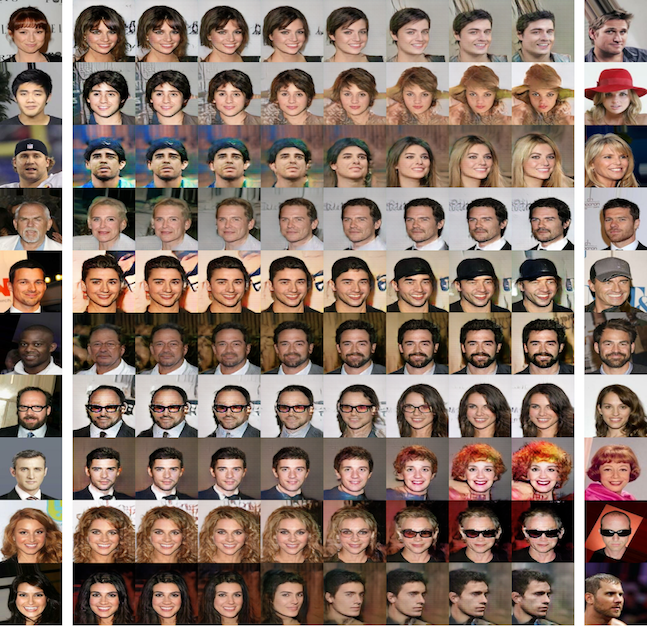}
    \caption{Interpolations from a BIGAN trained on the CelebA dataset.}
    \label{fig:celeba_inter}
\end{figure*}

\end{document}